\documentclass[11pt,a4paper]{article}
\usepackage[hyperref]{emnlp2020}
\usepackage{times}
\usepackage{latexsym}
\usepackage{dirtytalk}
\usepackage{array}
\usepackage{amsmath}
\usepackage{amssymb}
\usepackage{booktabs}
\usepackage{multirow}
\usepackage{graphicx}
\usepackage{gb4e}
\noautomath

\usepackage{microtype}

\aclfinalcopy % Uncomment this line for the final submission
\def\aclpaperid{3442} %  Enter the acl Paper ID here

\definecolor{yello}{HTML}{ffb677}
\definecolor{blu}{HTML}{005082}
\definecolor{purpl}{HTML}{726a95}
\definecolor{orang}{HTML}{ff9a76}
\definecolor{tealish}{HTML}{1aa6b7}

\newcommand{\blank}{$\rule{0.6cm}{0.15mm}$}

\newcommand{\mask}{\textsc{[mask]}}
\newcommand{\cls}{\textsc{[cls]}}
\newcommand{\sep}{\textsc{[sep]}}

\newcommand{\f}{\mathbb{F}}
\newcommand{\cc}{\mathcal{C}}
\newcommand{\tw}{\mathcal{T}}
\newcommand{\rp}{\mathcal{R}}
\newcommand{\up}{\mathcal{U}}

\title{Exploring BERT's Sensitivity to Lexical Cues using Tests from Semantic Priming}

\author{Kanishka Misra$^1$\hspace{0.5cm}Allyson Ettinger$^2$\hspace{0.5cm}Julia Taylor Rayz$^1$\\
  $^1$Department of Computer and Information Technology, Purdue University \\
  $^2$Department of Linguistics, University of Chicago\\
  $^1$\texttt{\{kmisra, jtaylor1\}@purdue.edu}, $^2$\texttt{aettinger@uchicago.edu}}

\date{}

\begin{document}
\maketitle
\begin{abstract}
  Models trained to estimate word probabilities in context have become ubiquitous in natural language processing. 
  How do these models use lexical cues in context to inform their word probabilities? To answer this question, we present a case study analyzing the pre-trained BERT model with tests informed by semantic priming. 
  Using English lexical stimuli that show priming in humans, we find that BERT too shows ``priming,"  predicting a word with greater probability when the context includes a related word versus an unrelated one. 
  This effect decreases as the amount of information provided by the context increases. Follow-up analysis shows BERT to be increasingly distracted by related prime words as context becomes more informative, assigning \emph{lower} probabilities to related words.
  Our findings highlight the importance of considering contextual constraint effects when studying word prediction in these models, and highlight possible parallels with human processing.
  % However, notably larger priming in certain relations suggests BERT might use cues that hold those relations with the \textit{target} more strongly than others.
  \end{abstract}

\section{Introduction}
% \todo{Transfer over from cogsci 2020 template}
The field of natural language processing (NLP) has recently seen a dramatic shift toward the use of language model (LM)-based pre-training \cite{howard2018universal, peters2018deep}---training based on estimating word probabilities in context---as a foundation for learning of a wide range of tasks. Leading this charge was the BERT model \cite{devlinBERTPretrainingDeep2019}, which is optimized in part to use context information to predict masked words. Because of the impressively strong performance of BERT and its successors \citep{yang2019xlnet, liu2019roberta, clark2020electra}, there has been increasing need for understanding how these types of models work, and what linguistic properties LM-based pre-training confers upon them. 

In this paper, we focus on the question of how BERT uses individual lexical relations to inform word probabilities in context. For example, if a word like \emph{airplane} is prepended to (\ref{1a}), to what extent does this increase the model's probability for the word \emph{pilot} in the blank position in (\ref{1b})?

\begin{exe}
  \ex
  \label{ex1}
  \begin{xlist}
      \ex\label{1a} I want to become a \blank.
      \ex\label{1b} \textit{airplane.} I want to become a \blank.
  \end{xlist}
\end{exe}

\noindent
This question is particularly relevant because human brains show a robust phenomenon of \emph{semantic priming} \citep{mcnamara2005semantic}, in which the presence of a word such as \say{airplane} will give rise to faster reactions to a related word like \say{pilot}. We explore whether the same lexical relations that show priming in humans will also be utilized by BERT to influence word predictions in context.

Our analysis includes three experiments.
First, we test BERT's sensitivity to single-word lexical cues for word prediction in context, using word pairs that show priming in humans, and testing for influence of contextual constraint. We find clear priming in BERT, but this effect is primarily localized to contexts that are relatively unconstraining. Next, we examine how BERT's use of these lexical cues varies depending on the type of lexical relation. We find that certain relations---particularly antonymy, synonymy, and category relations---evoke more sensitivity in BERT than others.  
Finally, we take a closer look at lexical cue dynamics in cases of high-constraint contexts, and we find that in such contexts we often see a phenomenon of ``distraction" rather than priming, such that related words actively demote probabilities of counterpart target words. 

Our paper has two main contributions. First, we introduce a methodology for fine-grained exploration of lexical cue sensitivity in predictive models, grounded in lexical relation phenomena observed in humans. Second, we apply these methods to shed light on word prediction dynamics of the BERT model. We discuss implications of these findings for considerations of contextual constraint, and for parallels with human processing. We release our datasets and code for further testing.\footnote{Data and code available at \url{https://github.com/kanishkamisra/emnlp-bert-priming}}

\section{BERT as a Semantic Priming Subject}
\label{sec2}

\subsection{Semantic Priming}
\label{sec1.1}
To study BERT's sensitivity to single-word cues in context, we draw on data from \emph{semantic priming} observed in humans. Semantic priming is an experimental phenomenon widely studied in psycholinguistics,  in which participants show a speedup in response to a word stimulus during language tasks when the response is preceded by a semantically related word as opposed to an unrelated one \cite{mcnamara2005semantic}.
Participants perform tasks like pronouncing a word out loud (``naming'') or deciding whether a given string is a word or not (``lexical decision''). The word to which the response is made is referred to as the \textit{target} and the preceding stimuli are called \textit{primes} (either related or unrelated). Levels of priming are evaluated based on participants’ response times (RT).
% For example, the RT for a target (e.g. \textsc{doctor}) is observed to be faster when it is preceded by a related prime (e.g. \textsc{nurse}) than an unrelated prime (e.g. \textsc{book}). This speedup in response is indicative of the relation between \textsc{doctor}, and \textsc{nurse}, and 
The magnitude of the speedup in RT provides information about the strength of the lexical relation in the context of the participants' cognitive system. The stimuli used in semantic priming experiments elicit responses caused by implicit processing within humans, which makes them an ideal intrinsic testing ground for studying models' quantification of word relations. Leveraging this fact, we take word pairs that show priming in humans, and use them to test BERT's sensitivity to lexical cues that have various types of relations.

\subsection{Extending Semantic Priming to BERT}

In humans, semantic priming occurs due to the presence of a lexical associate that affects the speed of response to a stimulus. Analogously, we are interested in learning how BERT's behavior (defined as a change in word probability) is affected by a lexical cue present in its input context. We define semantic priming in BERT as an increase in the model's expectation for a target word (or a lack thereof) in a given context in the presence of a semantically related word as compared to an unrelated one. Consider the following example:
% Following example (\ref{ex1}), modified and restated below as (\ref{ex2}):
\begin{exe}
    \ex
    \label{ex2}
    \begin{xlist}
        \ex\label{2a} I want to become a \blank.
        \ex\label{2b} \textit{airplane.} I want to become a \blank.
        \ex\label{2c} \textit{table.} I want to become a \blank.
    \end{xlist}
\end{exe}
\noindent
If the probability of the target word, \textit{pilot} is greater in (\ref{2b}) as compared to that in (\ref{2c}), then we interpret that the related word (\emph{airplane}) primes BERT more than the unrelated word (\emph{table}) does, for the target \emph{pilot} in the context (\ref{2a}). Such a test ensures that the only difference in BERT's output for the blank position in both cases is due to the swapping of the primes, allowing us to infer the degree to which BERT relies on single word cues to inform its probability for the target word. Importantly, our work here is not trying to simulate human semantic priming experiments directly---the structure of our tests is adapted for BERT's conventional usage by placing words in context, and thus deviates from standard word-level priming structure.

\subsection{Predictive Constraints of Target Contexts}
% Under what conditions does BERT allow a single related word in its context to affect its word predictions? 
We test how BERT's sensitivity to individual prime words varies based on contextual constraints. Consider the following example for target word \textit{key}:

\begin{exe}
\ex
\label{ex3}
\begin{xlist}
    \ex \label{3a} He lost the \blank yesterday.
    \ex \label{3b} She opened the door using a \blank.
\end{xlist}
\end{exe}

\noindent
In (\ref{3a}), the blank position can be any word that satisfies the semantic role \textsc{theme-of} for the event \textsc{lose}. 
The blank position is far more constrained in (\ref{3b}), which requires a word that satisfies the semantic role \textsc{instrument-of} for the event \textsc{unlock-door}---a set limited to items such as \textit{key, lock-pick,} or perhaps \textit{screwdriver}. As a result, the sentence in (\ref{3b}) is highly constraining towards predicting a word denoting these concepts or their relatives.

Focusing on how the constraint imposed by the context affects our notion of priming allows us to explore how much more information about the target word, \textit{key}, prepending a related word like \emph{lock} can provide in a high-constraint context such as (\ref{3b}), beyond words such as \say{open} and \say{door}. We can then compare priming behavior when \textit{lock} is prepended to (\ref{3a}), which imposes fewer constraints on the blank position.

Our focus on contextual constraints is in part motivated by studies that use sentence contexts of varying constraint to study priming in humans. In particular, \citet{schwanenflugelSemanticRelatednessScope1988} found low-constraint contexts to show wider scope of facilitation in lexical decision tasks, as compared to high-constraint ones, which only showed facilitations for the best completions (highest cloze probability). That is, low-constraint contexts produced enhanced facilitation effects in cases when the target word has low probability in the context. Taking this into account, when the context is highly constrained towards a particular completion, we expect BERT to show less sensitivity to the presence of an additional lexical cue, which may not provide significant information over and above that of the already constraining context. We hypothesize that in low-constraint contexts, because every word (including the target) is a low probability completion, BERT will be more sensitive to the addition of a single word in the context, thus showing greater priming effects in our testing framework.

\section{Related Work}
By focusing on the aforementioned considerations, and borrowing from the semantic priming paradigm, we build on a growing precedent of using psycholinguistics-inspired tests which focus on discovering the underlying mechanisms and linguistic competence of neural network
based models, and how closely they approximate language processing phenomena observed in humans.
For example, syntactic phenomena have been studied within recurrent neural network (RNN) LMs by supplying controlled, hand-crafted inputs to compare word probabilities in context across syntactically correct and anomalous instances \citep{futrelletal2019neural}. This methodology has been applied to study subject-verb agreement \citep{linzen2016assessing, gulordavaColorlessGreenRecurrent2018}, garden-path effects \citep{cogsci18vanSchijndel, frank2019interaction, futrelletal2019neural}, and filler-gap dependencies \citep{wilcox2018rnn}.
Deviating from prior work that has predominantly focused on investigating syntactic phenomena in LMs, \citet{ettinger2020bert} investigates BERT's semantic and pragmatic inference knowledge by using stimuli from N400 experiments \cite{kutas1980reading}. The findings suggested that BERT accurately attributes nouns to their hypernyms, but struggles in presence of negation, highlighting a limitation of LM-based training objectives.

\paragraph{Syntactic Priming in LMs}
\citet{prasadUsingPrimingUncover2019} draw on the syntactic priming paradigm---priming observed for sentence structure rather than word association---to investigate the ability of LMs to represent syntactic regularities. They define priming as adaptation to new stimuli by fine-tuning models on similarly structured sentences using the language model objective and investigating cumulative sentence surprisals before and after adaptation. In addition to focusing on a different type of priming, our work differs in operating directly on pre-trained BERT, without relying on any fine-tuning, which allows us to investigate the outcomes of the model's pre-training process itself. 

\paragraph{Mispriming in LMs}
Building upon work by \citet{petroni2019language}, which queries LMs by analyzing output over knowledge base queries recast as cloze questions, \citet{kassnerschutze2020negated} introduce the ``mispriming" probe, which shows BERT to be easily distracted by misprimes---words chosen to be prepended to cloze-like sentences. For instance, BERT-large predicts \textit{Cicero} as the completion in place of the correct answer, \textit{Plato}, when the previous query is modified to ``\textit{Cicero? Platonism is named after [MASK].}" While their setup is similar to the one discussed in this paper, our work differs methodologically in two ways: 1) we base our experiments on word pairs with clear, cognitively-based lexical relationships, for which we can explore fine-grained relation differences, and 2) we compare related to unrelated primes (rather than comparing primed to unprimed contexts, as do \citet{kassnerschutze2020negated}), thus keeping constant the prepending of a word, so as to target lexical relation effects more precisely. Furthermore, in the present work we are focused additionally on the effects of contextual constraint on BERT's lexical sensitivity during inference.

\section{Methods}
\label{methods}
\subsection{Model Investigated: BERT}
BERT \citep{devlinBERTPretrainingDeep2019} is a deep bidirectional transformer \citep{vaswani2017attention} network, trained on pairs of sentences. It is pre-trained on: (1) the Masked Language Model objective (predicting missing words in context), and (2) the Next Sentence Prediction objective (predicting whether the first sentence of the pair follows the second). 
% Unlike recurrent neural networks, it uses the transformer architecture \citep{vaswani2017attention} which enables it to represent each word as a function of the representations of all other words in the sentence using a mechanism known as self-attention, thus having the immediate context information of the word. 
We test on two variants: BERT-base (110M parameters) and BERT-large (340M parameters).

\subsection{Data}
We use the Semantic Priming Project (SPP) \citep{hutchison2013semantic} as our source of human priming experiment data. This resource has previously been used to evaluate word embedding models such as word2vec \citep{mikolov2013distributed} and GloVe \citep{pennington2014glove} by measuring the amount of variance in priming response times explained by cosine similarity between words as a predictor \citep{ettingerlinzen2016evaluating, augusteetal2017evaluation}. The SPP is a large collection of priming data for 768 subjects for 3322 triples, represented as ($\tw$, $\rp$, $\up$), where $\tw$ is the target word, and $\rp$ and $\up$ are the related and unrelated primes, respectively.
% There are 1661 target words which occur twice, once as \textit{first-associates}, and once as \textit{other-associates}, relative to their related prime word. 
% The SPP provides reaction time (RT)s for the $T, R$ and $T, U$ pairs with four methodological variations: Lexical Decision and Naming Tasks, each with two different Stimulus Onset Asynchronies (SOA; 200ms and 1200ms), which denote the time between the onset of the prime and the onset of the target. Shorter SOAs are understood to be associated with passive and automatic mechanisms, while longer SOAs reflect slower, more intentional processes while responding during priming tasks \cite{hutchison2001great}.
To enable fair comparison, we filter out target words that do not occur in BERT's vocabulary, as well as instances in which some of the RTs were missing, leaving us with 92\% of the total triples ($n$ = 3058).

\paragraph{Stimulus Construction}
% \label{construction}
In addition to the SPP triples, we introduce another component to accommodate the nature of the BERT model: a context, $\cc$, which is a naturally-occurring sentence originally containing the target word $\tw$, now with $\tw$ replaced by the ``\mask{}" token. We test the model's expectation for $\tw$ in the masked position when $\cc$ is preceded by a related prime $\rp$, as well as when it is preceded by an unrelated prime $\up$, denoted as $(\rp, \cc)$ and $(\up, \cc)$ respectively.
We choose to embed $\tw$ in $\cc$ in order to better simulate BERT's standard usage, given that the model is pre-trained to predict words in sentence contexts. We choose the contexts $\cc$ to be naturally-occurring sentences, since BERT is trained on well-formed sentences that affect its word level expectation. Our target contexts are sampled from the concatenation of the ROCstories Corpus \citep{mostafazadehetal2016corpus}, and the train and test sets used in the \say{Story Cloze Test} task \citep{mostafazadeh2017lsdsem}, primarily due to the simplistic nature of the sentences.

For our prime contexts, we experiment with two scenarios: \textbf{(a) \textsc{word}}: where the prime word, followed by a period, `.' is prepended to the target context, and \textbf{(b) \textsc{sentence}:} where a neutral context, \textit{``the next word is "} followed by the prime word and a `.', is prepended to the target context. We add the \cls{} and \sep{} tokens at the beginning and the end of each stimulus, respectively, following previous studies with a similar setup. Table \ref{tab:stimuli} shows full example items from these different settings. We limit to single word or neutral sentence contexts for our prime words because any naturalistic sentence containing $\rp$ would be different from that containing $\up$, thus adding imbalanced noise from the non-prime words. The context $\cc$ for the target, by contrast, will remain constant given that the target is constant (for any pair of primes).

\begin{table}[]
  \centering
  \small
  \begin{tabular}{@{}cl@{}}
  \toprule
  Scenario                   & Stimulus                                                                                                     \\ \midrule
  \multirow{2}{*}{\textsc{word}}     & \begin{tabular}[c]{@{}l@{}}\cls{} \textit{airplane.} I wanted to become \\ a \mask{}. \sep{}\end{tabular}                \\
                            & \begin{tabular}[c]{@{}l@{}}\cls{} \textit{table.} I wanted to become\\ a \mask{}. \sep{}\end{tabular}                    \\ \midrule
  \multirow{2}{*}{\textsc{sentence}} & \begin{tabular}[c]{@{}l@{}}\cls{} \textit{The next word is airplane.}\\I wanted to become a \mask{}. \sep{}\end{tabular} \\
                            & \begin{tabular}[c]{@{}l@{}}\cls{} \textit{The next word is table.}\\I wanted to become a \mask{}.  \sep{}\end{tabular}    \\ \bottomrule
  \end{tabular}
  \caption{Example Stimuli, with prime contexts in italics. Here, $\tw$ = \textit{pilot}, $\rp$ = \textit{airplane}, and $\up$ = \textit{table}.}
  \label{tab:stimuli}
  \vspace{-1em}
  \end{table}

  \paragraph{Contextual Constraints} We analyze BERT's reliance on single-word lexical cues (our primes) to inform its target word probabilities under various predictive constraints placed on the \mask{} token. To compute constraint of a context, we take the most expected words under BERT-base and BERT-large, and average their probabilities. This effectively represents how predictable the masked word is in the un-primed context. 
% \new
{Our notion of constraint is grounded in psycholinguistic studies examining effects of sentence contexts \cite{schwanenflugelSemanticRelatednessScope1988, federmeier1999rose}, which estimate sentence constraint based on the cloze probability of the most expected word in context. Mathematically, the constraint of a context $\cc$ is defined as:}
% \vspace{-1em}
% \begin{equation*}
%     \text{constraint} = \max_{x \in \mathcal{V}} \text{P}_{\text{BERT}}(\text{[MASK]} = x),
% \end{equation*}

% \begin{equation*}
% \label{constrainteqn}
% \small
\vspace{-1.5em}
\begin{multline*}
        \text{constraint}(\cc) = \\ \frac{1}{2}\sum_{m \in \{\textit{b}, \textit{l}\}} \max_{x \in \mathcal{V}} P_m(\mask = x \mid \cc), 
\end{multline*}

\noindent
% \new{
where $P_{m}$ represents the probability distribution for \mask{} in the output of the BERT model, either base (\emph{b}) or large (\emph{l}), and $x$ is a token belonging to BERT's vocabulary, $\mathcal{V}$.
Our proposed constraint scores are thus bounded by $[0, 1]$.
% }
We calculate the constraint for all sentences in our corpus that contain the target words, and group them into 10 equal bins of width 0.1 each, i.e, a constraint score of 0.38 would be in bin 4.
% \footnote{We experimented with bin sizes 2, 5, and 10 but only report results for 10 in the main paper.} 
Additionally, as a control, we also use a synthetic and unconstraining target context that we refer to as neutral\footnote{Our choice of neutral prime context follows \citet{schwanenflugelSemanticRelatednessScope1988}.}: \say{\cls{} the last word of this sentence is \mask{}. \sep{}}. This neutral context provides the lowest constraint, as it contains no information about what the masked target word can be---any word in BERT's vocabulary can fit in its \mask{} position. To make robust conclusions about the effect of constraint, we only sample triples that have at least one target context in each of the 10 bins. We faced polysemy issues for 72 target words, in which the sense of the target in the originally sampled $\cc$ did not fit the lexical relation with the primes---we manually corrected these by re-selecting appropriate contexts from the corpus. We could not resolve this issue for 28 items, which we discarded.
% \footnote{We will include results before and after correction in our supplementary materials.}
This further reduces the number of unique triples to 2112 (69\% of the valid instances), with each triple being associated with 11 (10 bins and a neutral context) stimuli.
% \new{
\paragraph{Constraint Scores and Entropy} While we follow psycholinguistic precedent in defining contextual constraint based on the highest-probability completion of a given context, another obvious candidate for defining contextual constraint would be the entropy of the probability distribution for the \mask{} token. In this setting, the entropy would quantify the amount of uncertainty about the \mask{} token  when conditioning on the context: low-constraint contexts would produce high uncertainty, and therefore a high entropy value, while high-constraint contexts would produce lower entropy values. To establish the consistency of our chosen constraint measure with an entropy-based definition of constraint, for every context ($\cc$) in our experiments we compute the entropy of the probability distribution on the \mask{} token, averaging the entropies from the two BERT models (\emph{b, l}):
\vspace{-0.7em}
\begin{multline*}
    \label{entropyconstraint}
    H_{\text{constraint}}(\cc) = \\ -\frac{1}{2}\sum_{m \in \{\textit{b}, \textit{l}\}}\sum_{x \in \mathcal{V}} P_m(x \mid \cc)\log{P_m(x \mid \cc)}
\end{multline*}
The Pearson correlation between our constraint measure and $H_{\text{constraint}}(\cc)$ is -0.89, indicating a strong empirical relationship between constraint measured as the probability of the best completion and entropy of the predicted distribution. 

\subsection{Measuring Priming in BERT}

We use \textbf{surprisal} as our measure of the model's expectation for $\tw$ in the given context. The surprisal of a language model denotes the level of \say{surprise} of the model for a word $w$, in context $\cc$: 

\begin{equation*}
    \text{\textit{Surp}}(w \mid \cc) = -\text{log}_2 P(w \mid h_{\cc}),    
\end{equation*}
where $h_{\cc}$ is the hidden state of the model 
% \ake{possibly just ``model''} 
for the context. 
% \new
{Surprisal is an effective linking hypothesis between language model probabilities and measures of human language processing. 
For instance, surprisal derived from n-gram and RNN LMs was shown to be a significant predictor of (1) self-paced reading times, a measure of cognitive load incurred during sentence comprehension in humans \citep{hale2001probabilistic, levy2008expectation, smith2013effect}; and (2) the amplitude of the N400 event related potential (ERP) \citep{frank2013word}, an electrical response that corresponds to lexical and semantic processing in human brains \citep{kutas1980reading}.}

In our experiments, we define the level of priming in BERT, which we call ``Facilitation", as:
\begin{equation*}
    \mathbb{F} = \text{\textit{Surp}}(\tw \mid \up, \cc) - \text{\textit{Surp}}(\tw \mid \rp, \cc).
\end{equation*}

\noindent
% \new
{Due to the setup of our stimuli, the difference in BERT's surprisals for the target word $\tw$ between the context pairs (related vs.~unrelated) quantifies the degree to which the model is influenced by one isolated prime word over the other. This can be considered analogous to the difference in human response times in the context of related versus unrelated primes, reflecting differing strengths of lexical association between the prime and target words.} If BERT is sensitive to the presence of a related prime, as humans are, such that $\rp$ primes the model to predict $\tw$ more than $\up$ does, then BERT should show less \say{surprise}---i.e., produce higher probability---for $\tw$ in the context $(\rp, \cc)$, than in $(\up, \cc)$. 
In such cases, $\f$ will be positive.

\section{Analysis and Results}
% We now describe our analysis and present results from our tests for priming within BERT. 
% We test facilitation effects in BERT and its relationship to sentence constraint using linear mixed-effects models \citep{baayen2008mixed} with random intercepts for target words.
To test for statistical significance between facilitation in BERT and the contextual constraints imposed by stimuli, we use a linear mixed-effects model with constraint scores as fixed effects and include random intercepts for target words.
The pre-trained BERT models were accessed using the Transformers library \cite{Wolf2019HuggingFacesTS}.

\begin{figure*}[!t]
\centering
\includegraphics[width=0.68\textwidth]{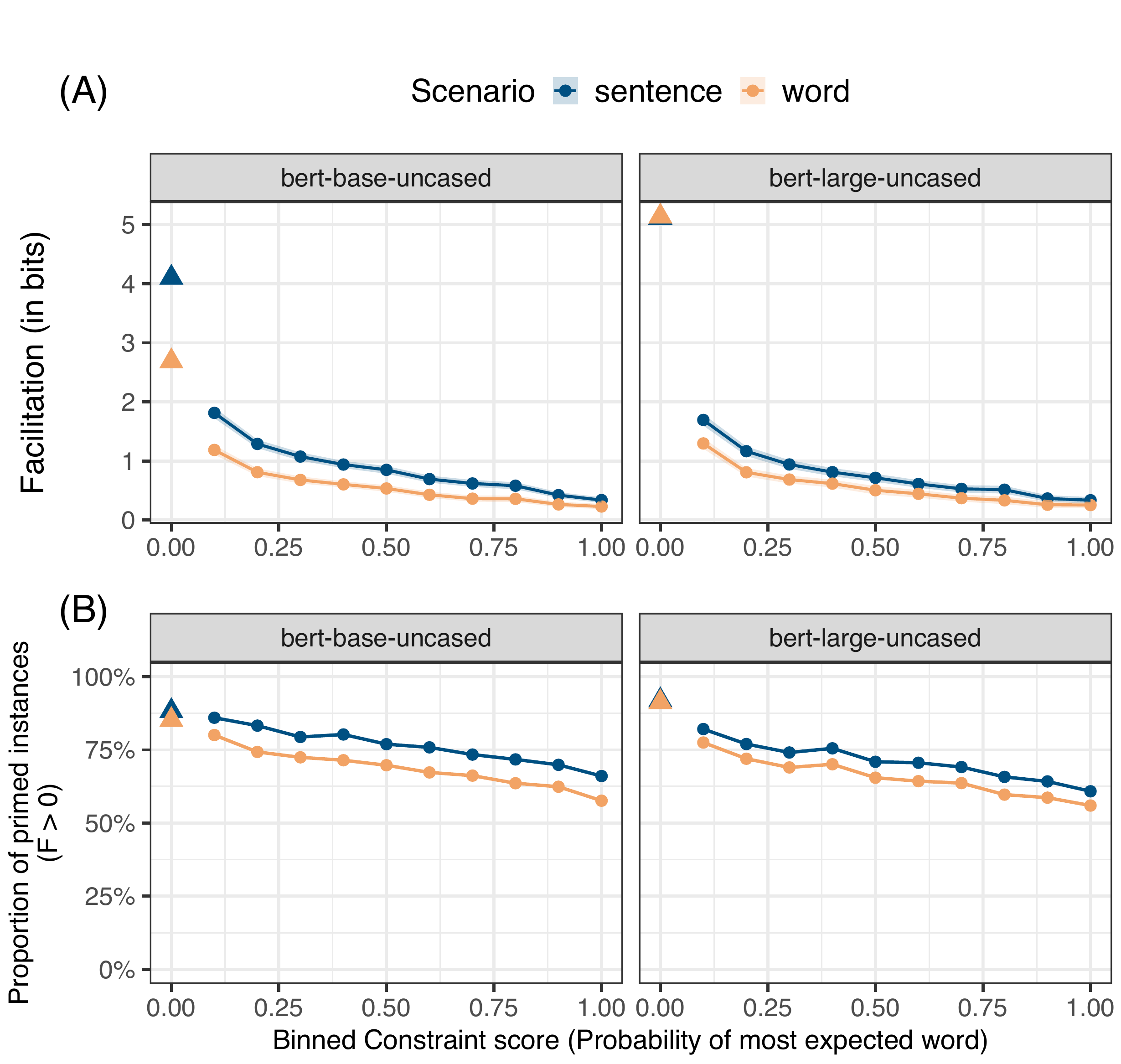}
\caption{Average facilitation (A) and proportion of primed instances, i.e., $\mathbb{F} > 0$ (B) vs. binned constraint score. Error bands represent 95\% confidence intervals. \textbf{Note:} Results for neutral contexts are shown separately as \textbf{triangles} (\textcolor{yello}{$\blacktriangle$}, \textcolor{blu}{$\blacktriangle$}), and do not correspond to a constraint score of 0.0 (actual constraint score = 0.02).}
\label{fig:overall}
% \vspace{-1em}
\end{figure*}

\subsection{How Facilitation is affected by Constraint}
\label{overall}
% We first compare facilitation as well as the number of instances where facilitation was observed, across our constraint bins by measuring the average facilitation per constraint bin, per model, per scenario. 
% The average facilitation results for both our models and for each prime context scenario, as well as the proportion of instances showing facilitation, are presented in Figure \ref{fig:overall}.
Figure \ref{fig:overall} shows the average facilitation effects and proportion of instances showing facilitation, for both models in each prime context setting.

% Overall, facilitation in BERT decreases as the predictive constraint placed on [MASK] in our stimuli increases, for both models across both our prime context scenarios (\textit{p} $<$ .001 in all cases\footnote{While we plot facilitation against binned constraint scores, the significance was derived using raw constraint value as a predictor in the linear mixed effects model.}). 
Overall, we find that priming in BERT decreases as the predictive constraint placed on the \mask{} position increases. This is evidenced by decrease in both the facilitation effect (\textit{p} $<$ .001 for both models in both scenarios),\footnote{While we plot facilitation against binned constraint scores, the significance was derived using raw constraint value as a predictor in the linear mixed effects model.} as well as the decrease in raw proportion of instances in which the facilitation was positive.
This indicates that the information provided by the related prime word (relative to the unrelated one) is increasingly outweighed by the information provided by the predictive constraints as the level of constraint increases. At lower levels of contextual constraint, BERT takes substantially more advantage of the lexical association of the prime word to predict the target word. This is particularly apparent in neutral contexts, where BERT receives almost no context information from non-prime words, and shows considerably larger facilitation.
Comparing settings with and without sentence context for the prime word, we see that BERT consistently shows greater facilitation effects when the prime context is a sentence rather than a single word, across every constraint bin (\textit{p} $<$ .001), with the exception of BERT-large for neutral contexts, where the magnitudes of the facilitation are the largest (as shown in Figure \ref{fig:overall}), but not significantly different between sentence and word prime contexts (\textit{t}(2111) = -0.3402, \textit{p} $=$ 0.6331).

\subsection{Facilitation across Lexical Relations}
% \label{}
We have established above that BERT's  predictions are sensitive to the addition of single related words in the context, particularly in contexts that are weakly constraining. In this section we investigate whether these sensitivity patterns are consistent across different types of lexical relations between the related prime and the target. We test priming effects for the 10 most frequent lexical relations annotated in the SPP, examples of which are shown in Table \ref{tab:relationexamples}.
As in section \ref{overall}, we test how facilitation changes with contextual constraints. 
% The subset of the SPP used in this work consists of 15 lexical relations---with two of them being \textit{unclassified} (\textit{n} = 471) and \textit{NA} (\textit{n} = 24), which are discarded in our analysis. 
\begin{figure*}[!t]
\centering
\includegraphics[width=\textwidth]{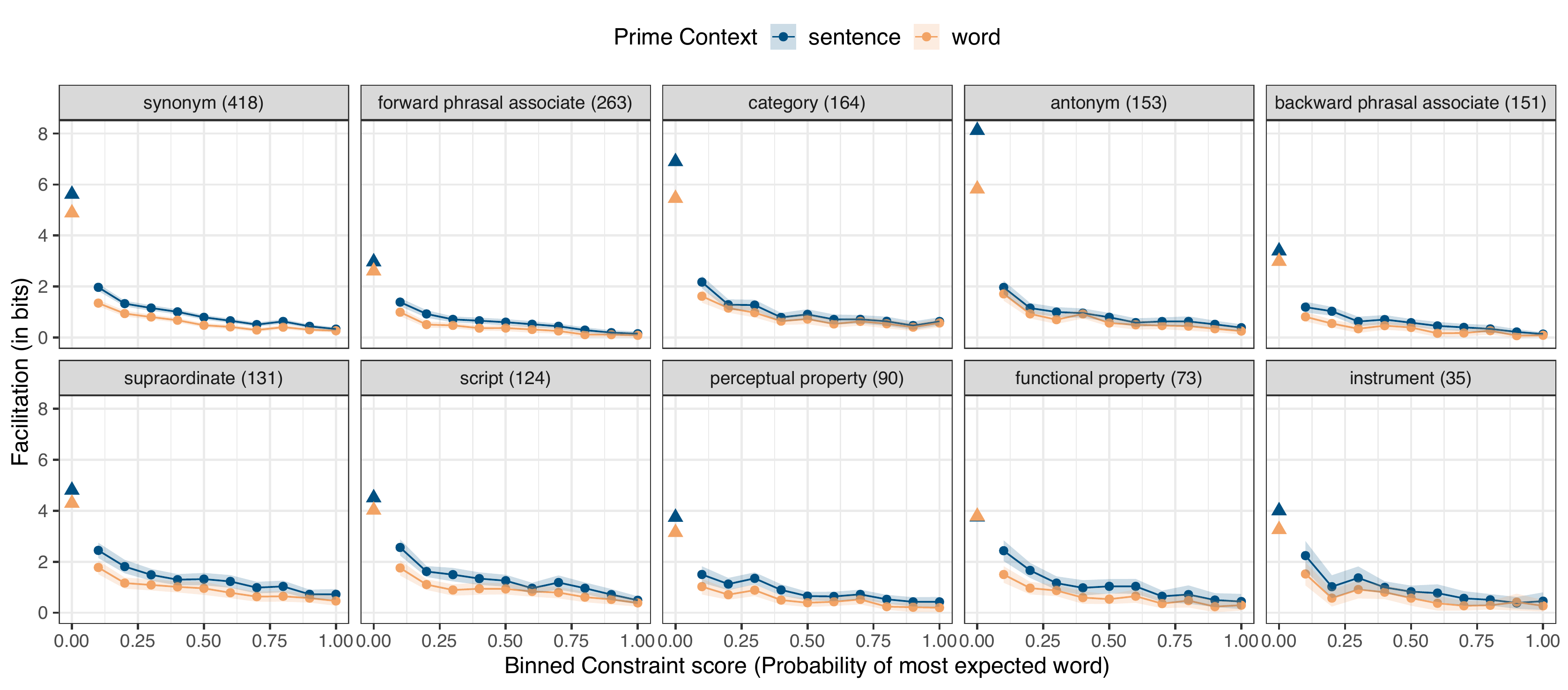}
\caption{Facilitation effects across top 10 lexical relations in our subset of SPP (averaged for BERT-base and BERT-large). Error bands represent 95\% confidence intervals. \textbf{Note:} Results for neutral contexts are shown separately as \textbf{triangles} (\textcolor{yello}{$\blacktriangle$}, \textcolor{blu}{$\blacktriangle$}), and do not correspond to a constraint score of 0.0 (actual constraint score = 0.02).}
\label{fig:relations}
\end{figure*}

\begin{table}[h]
% \vspace{-1.3em}
    \centering
    % \vskip 0.1in
        \small
\begin{tabular}{@{}lrl@{}}
\toprule
Relation & $n$ & $\tw, \rp$ \\ \midrule
Synonym & 418 & \textit{anger, fury} \\
Forward Phrasal Associate & 263 & \textit{ache, stomach} \\
Category & 164 & \textit{bed, sofa} \\
Antonym & 153 & \textit{deep, shallow} \\
Backward Phrasal Associate & 151 & \textit{cause, effect} \\
Supraordinate & 131 & \textit{spaghetti, pasta} \\
Script & 124 & \textit{judge, court} \\
Perceptual property & 90 & \textit{leaf, tree} \\
Functional property & 73 & \textit{bell, ring} \\
Instrument & 35 & \textit{bow, arrow} \\ \bottomrule
\end{tabular}
\caption{Top-10 relations within our subset of SPP.}
    \label{tab:relationexamples}
% \vspace{-2em}
\end{table}

Figure \ref{fig:relations} shows facilitation effects, averaged for BERT-base and BERT-large. We find facilitation effects across our subset of lexical relations to be consistent with the results in section \ref{overall}---facilitation decreases as the contextual constraint increases (\textit{p} $<$ .001 across all lexical relations and prime context scenarios). Among different lexical relations, we see the largest variation in BERT's sensitivity on the lower constraint items, which impose fewer restrictions on the identity of \mask{}. Synonymy, category, and antonymy relations show the most pronounced differences, with BERT showing considerably larger facilitation in the neutral context than for other relations. This suggests that BERT's word predictions in context may be more strongly attuned to relations of synonymy, category membership, and antonymy than to other lexical relations. 

\subsection{On Primes and Distractors}
\label{distractors}

\begin{figure}[!h]
% \vspace{-1em}
    \centering
    \includegraphics[width = 6cm]{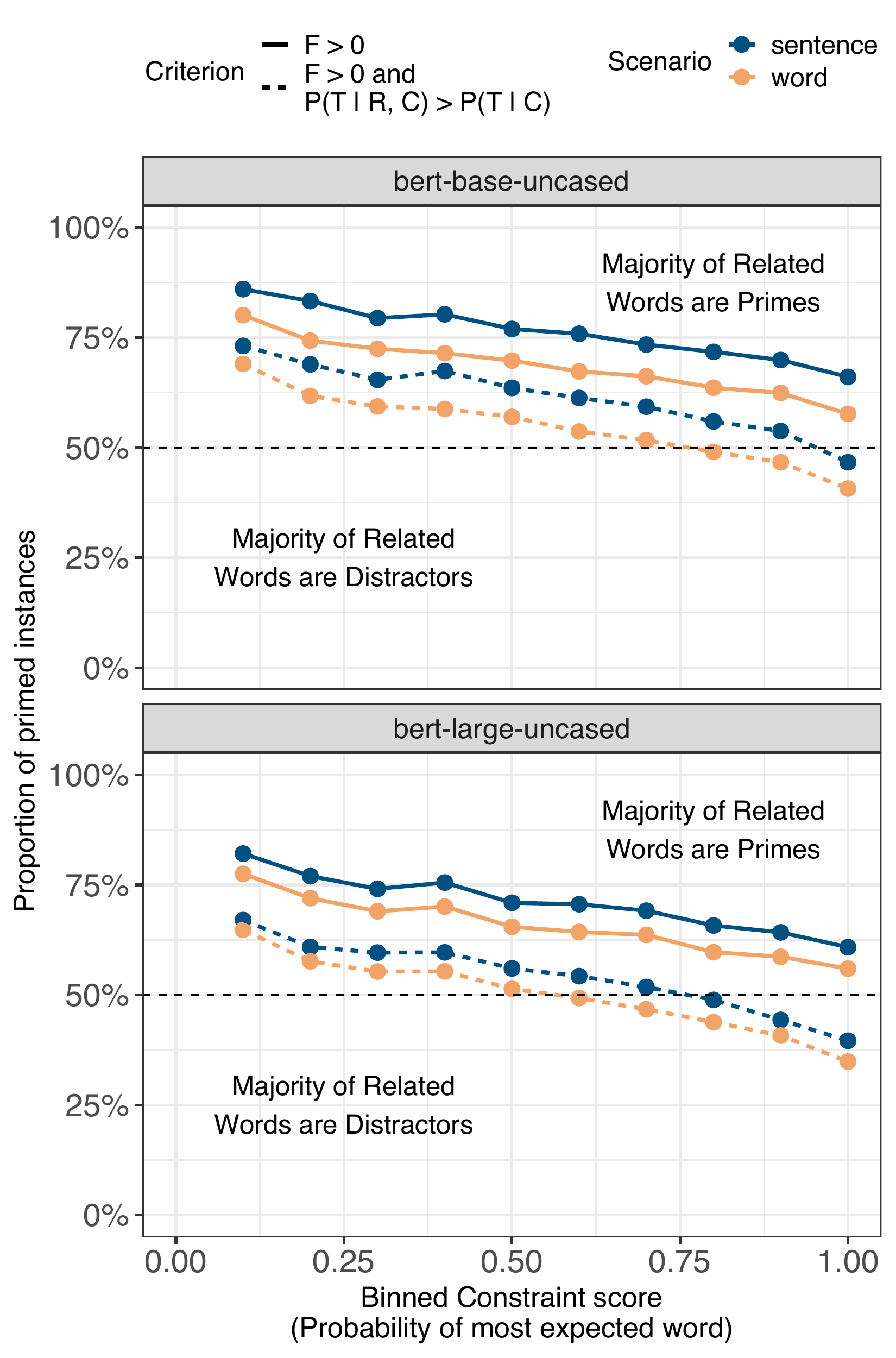}
    \caption{Proportion of primed instances under more (dashed) and less (solid) stringent priming criteria.}
    \label{fig:distractorfree}
    % \vspace{-0.7em}
\end{figure}

\begin{table*}[!t]
  \centering
  \small
  \begin{tabular}{@{}c|l|ll@{}}
  \toprule
  \multirow{2}{*}{\begin{tabular}[c]{@{}c@{}}Target\\ (Constraint)\end{tabular}} & \multirow{2}{*}{(\textit{Related / Unrelated}) Context} & \multicolumn{2}{c}{Top 5 Predicted Words (BERT-large probability)} \\ \cmidrule(l){3-4} 
   &  & \multicolumn{1}{c|}{Primed by Related} & \multicolumn{1}{c}{Primed by Unrelated} \\ \midrule
  \begin{tabular}[c]{@{}c@{}}\textit{bacon}\\ (0.89)\end{tabular} & \begin{tabular}[c]{@{}l@{}}(\textit{pork/meteorite}). she cooked up\\some eggs, \mask{}, and toast.\end{tabular} & \multicolumn{1}{l|}{\textit{\begin{tabular}[c]{@{}l@{}}eggs (0.20), potatoes (0.04),\\ tea (0.04), pancakes (0.04),\\ cheese (0.03)\end{tabular}}} & \textit{\begin{tabular}[c]{@{}l@{}}bacon (0.78), sausage (0.06),\\ ham (0.03), pancakes (0.02) \\ toast (0.02)\end{tabular}} \\ \midrule
  \begin{tabular}[c]{@{}c@{}}\textit{painting}\\ (0.75)\end{tabular} & \begin{tabular}[c]{@{}l@{}}(\textit{drawing/champagne}). dana was\\ a young artist who spent many\\ hours a day \mask{}.\end{tabular} & \multicolumn{1}{l|}{\textit{\begin{tabular}[c]{@{}l@{}}drawing (0.88), painting (0.10),\\ studying (\textless 0.01), writing (\textless 0.01),\\ practicing (\textless 0.01)\end{tabular}}} & \textit{\begin{tabular}[c]{@{}l@{}}painting (0.79), drawing (0.06),\\ working (0.03), studying (0.03),\\ teaching (0.01)\end{tabular}} \\ \bottomrule
  \end{tabular}
  \caption{Example high constraint instances that show ``distraction" rather than priming in BERT-large.}
  \label{tab:distraction}
  % \vspace{-1em}
  \end{table*}
  
  The preceding results show a decrease in number of primed instances as contextual constraint increases. This means that as the constraint imposed by the context increases, we see more instances in which the probability of the target word in presence of the related word is \emph{less} than that in presence of an unrelated word.
  For example, the first row of Table \ref{tab:distraction} shows an instance for a target, \textit{bacon}, with a constraint score of 0.89 (i.e., the 9$^\text{th}$ bin).
  Contrary to priming patterns observed in low-constraint contexts, the probability of \textit{bacon} is quite low when BERT is primed by \textit{pork}, and very high when the unrelated word, \textit{meteorite}, is the prime. 

  Here, the related prime acts as a \textbf{distractor},\footnote{We refer to it as a distractor since the target word is not the absolute correct completion for our contexts, since they are not factual like in \citet{kassnerschutze2020negated}.} similar to the \textbf{mispriming} reported in \citet{kassnerschutze2020negated}. Upon further investigation, we observe that the probability of the target word in presence of the related word is in fact also lower than that in an un-primed context, i.e., $P(\tw \mid \rp, \cc) <  P(\tw \mid \cc)$. The related word ``distracts" BERT, thereby reducing the probability of the target. To account for such cases, we make our criterion more stringent and count an instance as ``primed" if the facilitation is positive ($\mathbb{F} > 0$) \textbf{and} if the presence of the related word increases the probability of the target over that in the un-primed instance (P($\tw \mid \rp, \cc$) $>$ P($\tw \mid \cc$)). These changes are reflected in Figure \ref{fig:distractorfree}. 

  The proportion of facilitatory instances is now substantially lower with this more robust notion of priming, but it follows the same pattern observed when only facilitation score was considered. At higher constraint scores, the proportions fall under 50\%, giving us thresholds beyond which BERT shows more ``distraction" from related prime words than facilitation. For example, starting at the 8$^\text{th}$ constraint bin, BERT-base shows priming only for 49\% or fewer cases in the \textsc{word} prime context. 

% \vspace{-1.75em}
\paragraph{Qualitative Analysis} 
We examine specific instances of model predictions in order to shed further light on the factors that contribute to BERT's distraction (as opposed to priming) effects. Table \ref{tab:distraction} shows two examples in which we observe such distraction patterns in BERT. In the example with \textit{painting} as the target, we find BERT to show behavior akin to that discussed in \citet{kassnerschutze2020negated}. Here, the presence of a distractor (\textit{drawing}), one that fits as a completion in the \mask{} position, leads BERT to predict the distractor with greater probability than the target (\textit{painting)}. However, in the example with \textit{bacon} as the target, we observe a different kind of distraction: \textit{pork} cannot replace \textit{bacon} here as well as \textit{drawing} can replace \textit{painting} in the previous example, but 
  \textit{bacon} is still demoted in the probability distribution in favor of other foods related to \textit{pork}. By contrast, in both examples the unrelated primes resemble ``random misprimes" in \citet{kassnerschutze2020negated}: BERT isn't distracted by them---likely due to their degraded relevance to the context---and still predicts the target as the best completion.
  
  \section{General Discussion}
% \section{Discussion}
In the experiments above, we show that when using word pairs informed by human semantic priming, the BERT model is reliably sensitive to individual lexical cues in its context---\emph{if} the context is minimally constraining, such that there is little predictive information beyond that lexical cue.
% ---this pattern is consistent across diverse relation types---
As the predictive constraint applied by the context increases, BERT's level of sensitivity to a given lexical cue decreases. 
% Under lower contextual constraints, we find BERT to be primed for synonyms, antonyms, and category lexical relations considerably more than for other types of lexical relations. 
These results suggest that BERT uses lexical cues as needed: when informative sentence cues are available, single lexical items are of less value, and so they exert less influence on BERT's expectations for a masked word.

Examining patterns across different types of lexical relations, we find that this general effect of constraint holds across relation types, but synonym, category, and antonym relations elicit larger lexical sensitivities in BERT, as compared to other relations (when the context is unconstraining). 
This suggests that BERT has identified these relations---or the particular words that share these relations---as being more reliably predictive. This may be because words sharing these relations are simply more likely to co-occur during BERT's training, or BERT may have formed higher-order relational associations that inform these sensitivities. 

While we see that these priming-based lexical relations can have facilitatory effects on BERT's word predictions when the context is otherwise unconstraining, we see conversely that when the context \emph{is} constraining, prime words can actually have a ``distractor'' effect---actively demoting the target word in the probability distribution. This finding builds on recent evidence of BERT's sensitivity to such distractions when predicting completions to factual queries \citep{kassnerschutze2020negated}. We find in our analyses that the nature of this distraction depends critically on the interaction of contextual constraint and the strength of the lexical relation: when the context is unconstraining, the probability of a word is likely to be promoted by a related lexical item more than by an unrelated lexical item. If the context is constraining, a related lexical item may demote the probability of a target word in the predicted distribution, while an unrelated word is likely to have less impact. 

The effectiveness of human priming pairs in influencing BERT's lexical sensitivities, as well as the impact of contextual constraint on BERT's use of lexical context cues, suggest possible parallels with mechanisms in human language processing. Not only do humans show priming with the same word pairs that we show to impact BERT's predictions here, but like BERT, 
% shows an indifference to the presence of lexical cues in contexts that are already highly predictive of a single word, similarly, humans show priming only for the most predictable word in highly constraining sentences 
humans also show more limited semantic priming in constraining contexts, and wider scope of priming in low-constraint contexts \citep{schwanenflugelSemanticRelatednessScope1988}. 
% On the other hand, when prepended by a single word, low constraint contexts give rise to greater facilitation within BERT, akin to humans, who also show a wider scope of priming in low constraint settings. 
This suggests that the mechanisms that dictate BERT's lexical sensitivity may be optimized in a manner---or at least to an outcome---comparable to those underlying priming effects in humans. 

In practical terms, our results highlight the importance of contextual constraint in the dynamics of word prediction and information usage in the BERT model. Future work studying these dynamics should be mindful of this fact, as any observed prediction dynamics may change with the predictiveness of the context. This further emphasizes parallels with the study of human processing, as the predictive constraint of context has long been an important consideration and instrument in studying human sentence processing \citep{schwanenflugelSemanticRelatednessScope1988, schwanenflugel1991contextual, federmeier1999rose, mcfalls2002influence}. Our results show a similarly important role played by the amount of constraint imposed on a masked word during word probability estimation, which can lead to substantially different outcomes in behavioral analysis of pre-trained models.

\section{Conclusion and Future Work}

In this paper, we presented a framework to test how BERT
% ---a powerful language processing model that generates word predictions in context---
uses individual lexical relationships as cues for word prediction. Our framework is inspired by the psycholinguistic phenomenon of semantic priming, and our lexical cues are derived from a large collection of human priming experiments.

We examine the dynamics of BERT's word prediction in context, and relate its sensitivity towards lexical cues with contextual constraints and finer-grained lexical relations. Our findings establish the importance of considering predictive constraint effects of context in studies that behaviorally analyze language processing models, and highlight possible parallels with human processing.

The tests here are limited to the bidirectional masked language modeling framework used for training BERT, as opposed to autoregressive LM architectures such as RNNs, or GPT-2 \citep{radford2019language}. In future work it will be informative to establish whether different architectures and training objectives will produce differences in sensitivities towards contextual cues. Our paradigm can be extended by complementing our sampling procedure with hand-crafted templates of simple sentences that place all context to the left of target words. This will enable testing in the context of incremental language processing and help compare priming across various LM strategies.

\section{Acknowledgements}
We would like to thank the three anonymous reviewers for their helpful comments and suggestions. This work has also benefited from fruitful discussions with the members of the AKRaNLU lab at Purdue University, and the CompLing lab at the University of Chicago.

\bibliographystyle{acl_natbib}
\bibliography{anthology,emnlp2020, references}

\begin{thebibliography}{36}
\expandafter\ifx\csname natexlab\endcsname\relax\def\natexlab#1{#1}\fi

\bibitem[{Auguste et~al.(2017)Auguste, Rey, and
  Favre}]{augusteetal2017evaluation}
Jeremy Auguste, Arnaud Rey, and Benoit Favre. 2017.
\newblock Evaluation of word embeddings against cognitive processes: primed
  reaction times in lexical decision and naming tasks.
\newblock In \emph{Proceedings of the 2nd Workshop on Evaluating Vector Space
  Representations for {NLP}}, pages 21--26, Copenhagen, Denmark. Association
  for Computational Linguistics.

\bibitem[{Clark et~al.(2020)Clark, Luong, Le, and Manning}]{clark2020electra}
Kevin Clark, Minh-Thang Luong, Quoc~V. Le, and Christopher~D. Manning. 2020.
\newblock \href {https://openreview.net/forum?id=r1xMH1BtvB} {{ELECTRA:
  Pre-training Text Encoders as Discriminators Rather Than Generators}}.
\newblock In \emph{{International Conference on Learning Representations}}.

\bibitem[{Devlin et~al.(2019)Devlin, Chang, Lee, and
  Toutanova}]{devlinBERTPretrainingDeep2019}
Jacob Devlin, Ming-Wei Chang, Kenton Lee, and Kristina Toutanova. 2019.
\newblock {{BERT}}: {{Pre}}-training of {{Deep Bidirectional Transformers}} for
  {{Language Understanding}}.
\newblock In \emph{{Proceedings of NAACL-HLT 2019}}, pages 4171--4186,
  {Minneapolis, Minnesota}. {Association for Computational Linguistics}.

\bibitem[{Ettinger(2020)}]{ettinger2020bert}
Allyson Ettinger. 2020.
\newblock What {BERT} is not: Lessons from a new suite of psycholinguistic
  diagnostics for language models.
\newblock \emph{Transactions of the Association for Computational Linguistics},
  8:34--48.

\bibitem[{Ettinger and Linzen(2016)}]{ettingerlinzen2016evaluating}
Allyson Ettinger and Tal Linzen. 2016.
\newblock Evaluating vector space models using human semantic priming results.
\newblock In \emph{Proceedings of the 1st Workshop on Evaluating Vector-Space
  Representations for {NLP}}, pages 72--77, Berlin, Germany. Association for
  Computational Linguistics.

\bibitem[{Federmeier and Kutas(1999)}]{federmeier1999rose}
Kara~D Federmeier and Marta Kutas. 1999.
\newblock A rose by any other name: Long-term memory structure and sentence
  processing.
\newblock \emph{Journal of memory and Language}, 41(4):469--495.

\bibitem[{Frank and Hoeks(2019)}]{frank2019interaction}
Stefan~L Frank and John Hoeks. 2019.
\newblock The interaction between structure and meaning in sentence
  comprehension: Recurrent neural networks and reading times.
\newblock In \emph{CogSci 2019}, pages 337--343.

\bibitem[{Frank et~al.(2013)Frank, Otten, Galli, and Vigliocco}]{frank2013word}
Stefan~L Frank, Leun~J Otten, Giulia Galli, and Gabriella Vigliocco. 2013.
\newblock Word surprisal predicts n400 amplitude during reading.
\newblock In \emph{Proceedings of the 51st Annual Meeting of the Association
  for Computational Linguistics (Volume 2: Short Papers)}, pages 878--883.

\bibitem[{Futrell et~al.(2019)Futrell, Wilcox, Morita, Qian, Ballesteros, and
  Levy}]{futrelletal2019neural}
Richard Futrell, Ethan Wilcox, Takashi Morita, Peng Qian, Miguel Ballesteros,
  and Roger Levy. 2019.
\newblock Neural language models as psycholinguistic subjects: Representations
  of syntactic state.
\newblock In \emph{Proceedings of {NAACL-HLT 2019}}, pages 32--42, Minneapolis,
  Minnesota. Association for Computational Linguistics.

\bibitem[{Gulordava et~al.(2018)Gulordava, Bojanowski, Grave, Linzen, and
  Baroni}]{gulordavaColorlessGreenRecurrent2018}
Kristina Gulordava, Piotr Bojanowski, Edouard Grave, Tal Linzen, and Marco
  Baroni. 2018.
\newblock Colorless {{Green Recurrent Networks Dream Hierarchically}}.
\newblock In \emph{{Proceedings of NAACL-HLT} 2018}, pages 1195--1205, {New
  Orleans, Louisiana}. {Association for Computational Linguistics}.

\bibitem[{Hale(2001)}]{hale2001probabilistic}
John Hale. 2001.
\newblock \href {https://www.aclweb.org/anthology/N01-1021} {A probabilistic
  {E}arley parser as a psycholinguistic model}.
\newblock In \emph{Second Meeting of the North {A}merican Chapter of the
  Association for Computational Linguistics}.

\bibitem[{Howard and Ruder(2018)}]{howard2018universal}
Jeremy Howard and Sebastian Ruder. 2018.
\newblock Universal language model fine-tuning for text classification.
\newblock In \emph{Proceedings of the 56th Annual Meeting of the Association
  for Computational Linguistics (Volume 1: Long Papers)}, pages 328--339.

\bibitem[{Hutchison et~al.(2013)Hutchison, Balota, Neely, Cortese,
  Cohen-Shikora, Tse, Yap, Bengson, Niemeyer, and
  Buchanan}]{hutchison2013semantic}
Keith~A Hutchison, David~A Balota, James~H Neely, Michael~J Cortese, Emily~R
  Cohen-Shikora, Chi-Shing Tse, Melvin~J Yap, Jesse~J Bengson, Dale Niemeyer,
  and Erin Buchanan. 2013.
\newblock The semantic priming project.
\newblock \emph{Behavior Research Methods}, 45(4):1099--1114.

\bibitem[{Kassner and Sch{\"u}tze(2020)}]{kassnerschutze2020negated}
Nora Kassner and Hinrich Sch{\"u}tze. 2020.
\newblock \href {https://doi.org/10.18653/v1/2020.acl-main.698} {Negated and
  misprimed probes for pretrained language models: Birds can talk, but cannot
  fly}.
\newblock In \emph{Proceedings of the 58th Annual Meeting of the Association
  for Computational Linguistics}, pages 7811--7818, Online. Association for
  Computational Linguistics.

\bibitem[{Kutas and Hillyard(1980)}]{kutas1980reading}
Marta Kutas and Steven~A Hillyard. 1980.
\newblock Reading senseless sentences: Brain potentials reflect semantic
  incongruity.
\newblock \emph{Science}, 207(4427):203--205.

\bibitem[{Levy(2008)}]{levy2008expectation}
Roger Levy. 2008.
\newblock Expectation-based syntactic comprehension.
\newblock \emph{Cognition}, 106(3):1126--1177.

\bibitem[{Linzen et~al.(2016)Linzen, Dupoux, and
  Goldberg}]{linzen2016assessing}
Tal Linzen, Emmanuel Dupoux, and Yoav Goldberg. 2016.
\newblock Assessing the ability of lstms to learn syntax-sensitive
  dependencies.
\newblock \emph{Transactions of the Association for Computational Linguistics},
  4:521--535.

\bibitem[{Liu et~al.(2019)Liu, Ott, Goyal, Du, Joshi, Chen, Levy, Lewis,
  Zettlemoyer, and Stoyanov}]{liu2019roberta}
Yinhan Liu, Myle Ott, Naman Goyal, Jingfei Du, Mandar Joshi, Danqi Chen, Omer
  Levy, Mike Lewis, Luke Zettlemoyer, and Veselin Stoyanov. 2019.
\newblock {RoBERTa: A Robustly Optimized BERT Pretraining Approach}.
\newblock \emph{arXiv preprint arXiv:1907.11692}.

\bibitem[{McFalls and Schwanenflugel(2002)}]{mcfalls2002influence}
Elisabeth~L McFalls and Paula~J Schwanenflugel. 2002.
\newblock The influence of contextual constraints on recall for words within
  sentences.
\newblock \emph{American Journal of Psychology}, 115(1):67--88.

\bibitem[{McNamara(2005)}]{mcnamara2005semantic}
Timothy~P McNamara. 2005.
\newblock \emph{Semantic priming: Perspectives from memory and word
  recognition}.
\newblock Psychology Press.

\bibitem[{Mikolov et~al.(2013)Mikolov, Sutskever, Chen, Corrado, and
  Dean}]{mikolov2013distributed}
Tomas Mikolov, Ilya Sutskever, Kai Chen, Greg~S Corrado, and Jeff Dean. 2013.
\newblock Distributed representations of words and phrases and their
  compositionality.
\newblock In \emph{Advances in neural information processing systems}, pages
  3111--3119.

\bibitem[{Mostafazadeh et~al.(2016)Mostafazadeh, Chambers, He, Parikh, Batra,
  Vanderwende, Kohli, and Allen}]{mostafazadehetal2016corpus}
Nasrin Mostafazadeh, Nathanael Chambers, Xiaodong He, Devi Parikh, Dhruv Batra,
  Lucy Vanderwende, Pushmeet Kohli, and James Allen. 2016.
\newblock A corpus and cloze evaluation for deeper understanding of commonsense
  stories.
\newblock In \emph{Proceedings of {NAACL-HLT} 2016}, pages 839--849, San Diego,
  California. Association for Computational Linguistics.

\bibitem[{Mostafazadeh et~al.(2017)Mostafazadeh, Roth, Louis, Chambers, and
  Allen}]{mostafazadeh2017lsdsem}
Nasrin Mostafazadeh, Michael Roth, Annie Louis, Nathanael Chambers, and James
  Allen. 2017.
\newblock {LSDSem 2017 Shared Task: The Story Cloze Test}.
\newblock In \emph{Proceedings of the 2nd Workshop on Linking Models of
  Lexical, Sentential and Discourse-level Semantics}, pages 46--51.

\bibitem[{Pennington et~al.(2014)Pennington, Socher, and
  Manning}]{pennington2014glove}
Jeffrey Pennington, Richard Socher, and Christopher Manning. 2014.
\newblock Glove: Global vectors for word representation.
\newblock In \emph{Proceedings of {EMNLP 2014}}, pages 1532--1543.

\bibitem[{Peters et~al.(2018)Peters, Neumann, Iyyer, Gardner, Clark, Lee, and
  Zettlemoyer}]{peters2018deep}
Matthew Peters, Mark Neumann, Mohit Iyyer, Matt Gardner, Christopher Clark,
  Kenton Lee, and Luke Zettlemoyer. 2018.
\newblock Deep contextualized word representations.
\newblock In \emph{Proceedings of the 2018 Conference of the North American
  Chapter of the Association for Computational Linguistics: Human Language
  Technologies, Volume 1 (Long Papers)}, pages 2227--2237.

\bibitem[{Petroni et~al.(2019)Petroni, Rockt{\"a}schel, Riedel, Lewis, Bakhtin,
  Wu, and Miller}]{petroni2019language}
Fabio Petroni, Tim Rockt{\"a}schel, Sebastian Riedel, Patrick Lewis, Anton
  Bakhtin, Yuxiang Wu, and Alexander Miller. 2019.
\newblock Language models as knowledge bases?
\newblock In \emph{Proceedings of the 2019 Conference on Empirical Methods in
  Natural Language Processing and the 9th International Joint Conference on
  Natural Language Processing (EMNLP-IJCNLP)}, pages 2463--2473.

\bibitem[{Prasad et~al.(2019)Prasad, {van Schijndel}, and
  Linzen}]{prasadUsingPrimingUncover2019}
Grusha Prasad, Marten {van Schijndel}, and Tal Linzen. 2019.
\newblock Using {{Priming}} to {{Uncover}} the {{Organization}} of {{Syntactic
  Representations}} in {{Neural Language Models}}.
\newblock In \emph{Proceedings of the 23rd {{Conference}} on {{Computational
  Natural Language Learning}} ({{CoNLL}})}, pages 66--76, {Hong Kong, China}.
  {Association for Computational Linguistics}.

\bibitem[{Radford et~al.(2019)Radford, Wu, Child, Luan, Amodei, and
  Sutskever}]{radford2019language}
Alec Radford, Jeff Wu, Rewon Child, David Luan, Dario Amodei, and Ilya
  Sutskever. 2019.
\newblock Language models are unsupervised multitask learners.

\bibitem[{van Schijndel and Linzen(2018)}]{cogsci18vanSchijndel}
Marten van Schijndel and Tal Linzen. 2018.
\newblock Modeling garden path effects without explicit hierarchical syntax.
\newblock In \emph{CogSci 2018}, pages 2603--2608.

\bibitem[{Schwanenflugel(1991)}]{schwanenflugel1991contextual}
Paula~J Schwanenflugel. 1991.
\newblock Contextual constraint and lexical processing.
\newblock In \emph{Advances in psychology}, volume~77, pages 23--45. Elsevier.

\bibitem[{Schwanenflugel and
  LaCount(1988)}]{schwanenflugelSemanticRelatednessScope1988}
Paula~J. Schwanenflugel and Kathy~L. LaCount. 1988.
\newblock Semantic relatedness and the scope of facilitation for upcoming words
  in sentences.
\newblock \emph{Journal of Experimental Psychology: Learning, Memory, and
  Cognition}, 14(2):344--354.

\bibitem[{Smith and Levy(2013)}]{smith2013effect}
Nathaniel~J Smith and Roger Levy. 2013.
\newblock The effect of word predictability on reading time is logarithmic.
\newblock \emph{Cognition}, 128(3):302--319.

\bibitem[{Vaswani et~al.(2017)Vaswani, Shazeer, Parmar, Uszkoreit, Jones,
  Gomez, Kaiser, and Polosukhin}]{vaswani2017attention}
Ashish Vaswani, Noam Shazeer, Niki Parmar, Jakob Uszkoreit, Llion Jones,
  Aidan~N Gomez, {\L}ukasz Kaiser, and Illia Polosukhin. 2017.
\newblock Attention is all you need.
\newblock In \emph{Advances in neural information processing systems}, pages
  5998--6008.

\bibitem[{Wilcox et~al.(2018)Wilcox, Levy, Morita, and Futrell}]{wilcox2018rnn}
Ethan Wilcox, Roger Levy, Takashi Morita, and Richard Futrell. 2018.
\newblock What do rnn language models learn about filler--gap dependencies?
\newblock In \emph{Proceedings of the 2018 EMNLP Workshop BlackboxNLP:
  Analyzing and Interpreting Neural Networks for NLP}, pages 211--221.

\bibitem[{Wolf et~al.(2019)Wolf, Debut, Sanh, Chaumond, Delangue, Moi, Cistac,
  Rault, Louf, Funtowicz, and Brew}]{Wolf2019HuggingFacesTS}
Thomas Wolf, Lysandre Debut, Victor Sanh, Julien Chaumond, Clement Delangue,
  Anthony Moi, Pierric Cistac, Tim Rault, R'emi Louf, Morgan Funtowicz, and
  Jamie Brew. 2019.
\newblock Huggingface's transformers: State-of-the-art natural language
  processing.
\newblock \emph{ArXiv}, abs/1910.03771.

\bibitem[{Yang et~al.(2019)Yang, Dai, Yang, Carbonell, Salakhutdinov, and
  Le}]{yang2019xlnet}
Zhilin Yang, Zihang Dai, Yiming Yang, Jaime Carbonell, Russ~R Salakhutdinov,
  and Quoc~V Le. 2019.
\newblock {XLNet: Generalized autoregressive pretraining for language
  understanding}.
\newblock In \emph{Advances in Neural Information Processing Systems}, pages
  5754--5764.

\end{thebibliography}

\end{document}